\documentclass{ifacconf}
\usepackage{natbib}
\usepackage{booktabs}
\usepackage{amsmath}
\usepackage{amsfonts}
\usepackage{graphicx}
\usepackage[activate={expansion,lh=false}, protrusion=false]{microtype}
\usepackage{url,xcolor}

\usepackage[acronym]{glossaries}
\newacronym{svm}{SVM}{Support Vector Machine}
\newacronym{dt}{DT}{Decision Tree}
\newacronym{rf}{RF}{Random Forest}
\glsdisablehyper

\begin{document}
\begin{frontmatter}

\title{SensHRPS: Sensing Comfortable Human-Robot Proxemics and Personal Space With Eye-Tracking} 

%\thanks[footnoteinfo]{Sponsor and financial support acknowledgment goes here. Paper titles should be written in uppercase and lowercase letters, not all uppercase.}

\author[RPTU]{Nadezhda Kushina} 
\author[DFKI]{Ko Watanabe} 
\author[RPTU]{Aarthi Kannan}
\author[RPTU]{Ashita Ashok}
\author[DFKI]{Andreas Dengel}
\author[RPTU]{Karsten Berns}

\address[RPTU]{RPTU Kaiserslautern-Landau, Erwin-Schrödinger-Straße 52, 67663 Kaiserslautern Germany (e-mail: kushina@rptu.de).}
\address[DFKI]{German Research Center of Artificial Intelligence,  Trippstadter Str. 122, 67663 Kaiserslautern Germany (e-mail: ko.watanabe@dfki.de)}

 %Regular Papers: Must have a minimum length of 4 pages and may be submitted with a manuscript of up to 8 pages for review. Accepted papers must have a final version of no more than 6 pages. Note that no overpage (more than 6) is allowed in the final version. These papers will be published in the Congress proceedings at IFAC-PapersOnLine (POL).
%
% Abstract of 50--100 words
\begin{abstract}
Social robots must adjust to human proxemic norms to ensure user comfort and engagement. While prior research demonstrates that eye-tracking features reliably estimate comfort in human-human interactions, their applicability to interactions with humanoid robots remains unexplored. In this study, we investigate user comfort with the robot "Ameca" across four experimentally controlled distances (0.5 m to 2.0 m) using mobile eye-tracking and subjective reporting (N=19). We evaluate multiple machine learning and deep learning models to estimate comfort based on gaze features. Contrary to previous human-human studies where Transformer models excelled, a Decision Tree classifier achieved the highest performance (F1-score = 0.73), with minimum pupil diameter identified as the most critical predictor. These findings suggest that physiological comfort thresholds in human-robot interaction differ from human-human dynamics and can be effectively modeled using interpretable logic. 
\end{abstract}


\end{frontmatter}
%===============================================================================

\section{Introduction}

%Why proxemics matters in HRI
In human-robot interaction (HRI), a robot’s nonverbal cues often determine the emotional resonance of an interaction more than the spoken content. Even if a robot delivers a benign phrase, behavioral misalignments, such as a fixed stare or intrusive proximity, can transform a standard exchange into an unnerving experience. Therefore, robots must adhere to human conversational norms to ensure user satisfaction and continued engagement \citep{henschel2020social, hedayati2023should}. Personal space is particularly dynamic and context-sensitive; violations induce discomfort and stress, making proxemic alignment a foundational requirement for effective HRI \citep{watanabe_sensps_2025, Petrak-proxemy-aware-2019}.

%Why humanoid robots introduce unique challenges
Highly anthropomorphic robot designs often trigger the uncanny valley effect, increasing the interpersonal distance required for user comfort \citep{berns2024you, lanfranchi2023estimation}.  Robot attributes such as appearance and height further influence these preferences \citep{walters2009empirical}. These factors are particularly relevant for tall social humanoid robots (SHRs) (see Research, Social \& Entertainment humanoids\footnote{https://www.merphi.se/download-robotic-poster/}), such as the Gen 1 Ameca robot from Engineered Arts\footnote{https://engineeredarts.com/robots/ameca}, whose physical form may amplify users' distance preferences.

%current gap
Research in human-robot proxemics (HRP) provides valuable insights but also reveals inconsistencies in how distance shapes interaction quality. While HRP has been examined across various contexts, results vary significantly based on experimental setup \citep{samarakoon2022review}. Notably, prior work has not examined how interlocutors experience comfort or discomfort with a tall, highly anthropomorphic, yet immobile humanoid robot at fixed and systematically varied distances. 

%How do studies study comfort?
Furthermore, methodologies for assessing comfort typically rely on explicit, subjective measures, such as participant-chosen distances or static questionnaires. While adaptive proxemic systems increasingly account for user posture, behavior, and traits \citep{samarakoon2022review}, the use of implicit physiological signals remains limited. \cite{watanabe_sensps_2025} recently demonstrated that gaze-based features reliably predict comfort in human–human interactions across controlled distances. However, it remains unclear whether these signals generalize to interactions with highly anthropomorphic humanoid robots, whose appearance, social presence, and immobility may alter comfort dynamics.
 \begin{figure}[t!]
  \centering
  \includegraphics[width=0.45\textwidth]{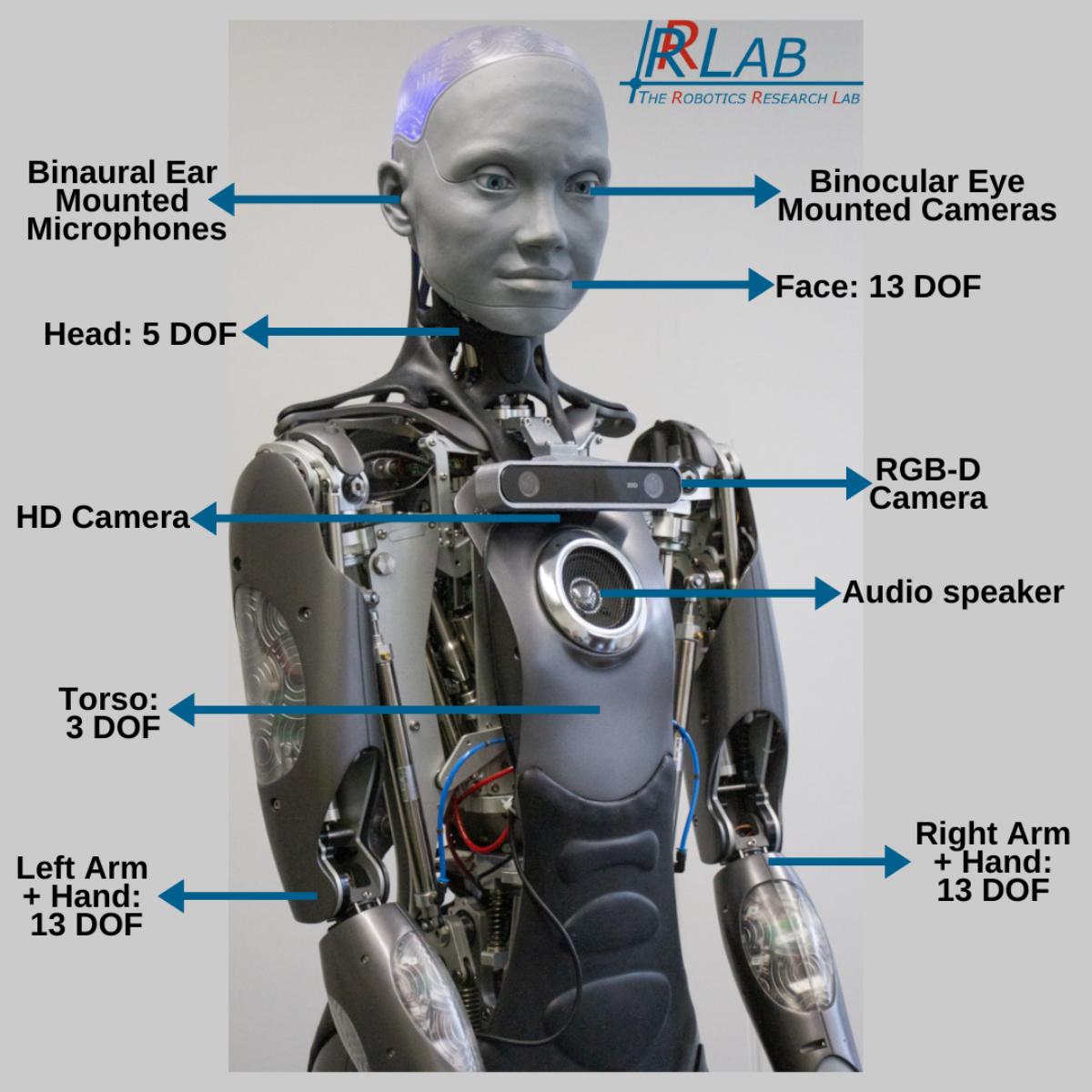}
  \caption{The social robot deployed in this study is ``Ameca''. Anthropomorphic Robot made by Engineered Arts.}
  \label{fig:communication_robot}
\end{figure}
To address this gap, we investigate how users experience robot Ameca (see Fig. \ref{fig:communication_robot}) at fixed distances during unconstrained small talk while wearing eye-tracking glasses. We utilize small talk because it facilitates human-robot communication \citep{babel_small_2021,ashok2024chit}. By experimentally controlling distance, we aim to correlate specific gaze metrics with user comfort levels in real-time. Our goal is to identify objective physiological thresholds for discomfort that can inform future adaptive robot behaviors. This leads to the research questions:

\begin{itemize}
\item {\textbf{RQ1}}: Can we estimate human-robot proxemics and personal space using gaze features?
\item {\textbf{RQ2}}: Which gaze features enable estimation of user comfort during HRI with highly anthropomorphic SHR?
\end{itemize}

% Literature highlights gaze behavior as a critical lens for understanding interpersonal distance in HRI \citep{mishra2024face}. 
% Prior work shows that mutual gaze can increase physical distancing in accordance with compensation models of interpersonal behavior and that this effect becomes stronger when the robot is perceived as unlikeable \cite{mumm2011human}. 

% These findings indicate that eye behavior may serve as an implicit measure of perceived social closeness and comfort around robots, motivating the following hypothesis:
%  \begin{itemize}
%  \item {\textbf{Hypothesis}}: Interlocutors will maintain less eye contact when their distance to the highly anthropomorphic SHR decreases.
%  \end{itemize}

% To our knowledge, no prior work has systematically measured comfort across controlled distances with a tall humanoid robot during natural LLM-driven conversation using mobile eye-tracking, leaving a gap in understanding how comfort evolves when distance increments are experimentally controlled. 
\section{Related Work}
\subsection{Personal Space and Comfort}
%what is personal space
Individuals tend to perceive the area around them as their personal space. The violation of this ``invisible zone'' causes negative feelings, such as threat and discomfort \citep{miller_more_2021}.While researchers distinguish between interpersonal space (human–human) and peripersonal space (human–environment) \citep{coello_interrelation_2021}, HRI blurs these lines.  Physical robots trigger a distinct cognitive frame of mind in which users experience them as real, co-present agents, creating stronger expectations for appropriate interpersonal spacing and proxemic awareness, especially for social humanoids, which are perceived as active participants in interpersonal space \citep{mutlu_virtual_2020, Petrak-proxemy-aware-2019}.

%The distances between humans are various for contexts and kinds of relationship, and into several zones, such as: Close intimate zone (0.15 to 0.45 m) – Personal zone (0.45 to 1.2 m) – Social zone (1.2 to 3.6 m) \citep{babel_small_2021}. Keeping the correct distance results in higher comfort. The appropriate distance is shaped by verbal and non-verbal communication, such as the choice of the topic and eye contact \citep{babel_small_2021}. Comfort itself has been conceptualized as a uni-dimensional evaluation ranging from extremely uncomfortable to extremely comfortable, defined as the approval or disapproval of an interaction situation and the resulting desire to maintain or withdraw from it \citep{redondo2020can}. Understanding and sensing comfort is considered essential for future adaptive robots, especially because violations of social norms can significantly influence perceptions over extended interactions \citep{redondo2024comfortability}, and distance expectations are dynamic and can be easily violated \citep{mutlu_virtual_2020}.

%When shifting focus from humans to robots, proxemic patterns diverge. 
% ( ~230–240 cm against ~280–300 cm). \cite{pazhoohi_give_2023} 

\subsection{Proxemics in HRI}
%proxemics with a robot and factors influencing it
 %Proxemic preferences in HRI are influenced by three primary vectors: \begin{itemize} \item \textbf{Individual Characteristics:} Cultural background, age, and individual personality traits \citep{babel_small_2021}. \item \textbf{Robot Characteristics:} Factors such as whether the robot is passive or approaching, its facial expressions, gaze behavior, size, and appearance \citep{samarakoon2022review, yamanaka_characteristics_2024}. \item \textbf{Context:} Familiarity, task type, and environmental setting \citep{babel_small_2021}.  \end{itemize}

Recent reviews indicate that proxemic preferences remain highly context-dependent and sensitive to robot morphology \citep{yamanaka_characteristics_2024, samarakoon2022review}.

Robots elicit stronger reactions when invading personal space and result in lower overall comfort. Preferred distances vary substantially depending on robot size, movement direction, facial expressivity, and social role, with taller or more human-like robots often eliciting greater preferred spacing \cite{babel_small_2021, yamanaka_characteristics_2024}. The existing literature is characterized by significant methodological heterogeneity in tasks and measurement techniques \citep{samarakoon2022review}. Although these studies provide important descriptive accounts of HRP, their heterogeneity illustrates that proxemic preferences are highly context dependent, making it difficult to generalize comfortable distances. 

\begin{table*}[t!]
\centering
\caption{Overview of pupil and gaze-related features.}
\renewcommand{\arraystretch}{1.2}
\begin{tabular}{p{4.8cm} p{3.5cm} p{8.5cm}}
\hline
\textbf{Category} & \textbf{Feature} & \textbf{Description} \\
\hline

Basic Pupil Position and Size & \texttt{diameter} &
Pupil diameter in pixels; can reflect alertness or cognitive load. \\

& \texttt{diameter\_3d} &
3D pupil diameter in millimeters; provides a more precise measure of pupil size. \\[0.2cm]

Ellipse Fitting Features & \texttt{ellipse\_axis\_a}, \texttt{ellipse\_axis\_b} &
Major and minor axes of the ellipse; characterize the pupil shape. \\[0.2cm]

3D Geometric Features & \texttt{circle\_3d\_center\_x}, \texttt{circle\_3d\_center\_y}, \texttt{circle\_3d\_center\_z} &
3D center coordinates of the pupil circle on the eyeball surface. \\[0.2cm]

Angular Features &
\texttt{theta}, \texttt{phi} &
Angular coordinates in a spherical coordinate system; compact representation of gaze direction. \\[0.2cm]

Temporal and Identifier Features &
\texttt{pupil\_timestamp} &
Timestamp of the pupil detection, used for time-series analysis and synchronization. \\

& \texttt{world\_index} &
Index of the corresponding frame in the world (scene) camera, used for synchronization. \\

& \texttt{eye\_id} &
Eye identifier (0 = left eye, 1 = right eye) for distinguishing and comparing both eyes. \\[0.2cm]

Confidence Features &
\texttt{confidence} &
Detection confidence of the pupil (already filtered at $\geq 0.8$). \\
\hline
\end{tabular}
\label{tab:pupil_features}
\end{table*}

\subsection{Physiological Comfort Estimation}
To move beyond subjective self-reporting, we look to physiological metrics. Eye-tracking correlates with transient states such as cognitive load \citep{melo_automatic_2025}, self-confidence \citep{Bhatt_confidence_2024}, and comfort \citep{watanabe_sensps_2025}. In particular, \cite{watanabe_sensps_2025} in their study ``SensPS'' used sensors to determine whether physiological data correlate with comfort during proxemic violations. In the experiment, ten participants wore eye-tracking glasses and a wristband while conversing with an experimenter at four distances ranging from 2 meters to 0.5 meters. The comfortable distance was categorized into two classes based on responses to a Likert-scale  6-item questionnaire \citep{watanabe_sensps_2025}. Participants rated their level of comfort on a scale from 1 to 10; 10 indicated comfortable, and any score below 10 indicated uncomfortable. Of all the collected physiological data (heart rate, electrodermal activity, and eye gaze), eye tracking was the most robust predictor of comfort. This was especially true for 3D gaze-point skewness and entropy, and for pupil diameter. Transformer model produced the best results, with an F1-score of 0.87. The study demonstrated that eye movements and pupil diameter are highly sensitive to changes in distance and strongly correlate with discomfort at close range (0.5 m). The findings support adaptive proxemics systems that modulate distance in response to gaze behavior.

In HRI, however, eye tracking is uncommon. For instance, \cite{hostettler2024realtimeadaptiveindustrialrobots} used pupil diameter to trigger adaptive behaviors in industrial settings. No prior work has measured comfort using mobile eye-tracking with a humanoid during natural language interaction. We posit that the mechanisms identified by ``SensPS'' \citep{watanabe_sensps_2025} may not directly transfer to HRI due to the "intentional stance" users adopt toward robots. Therefore, we empirically test these gaze features to determine their validity as objective comfort estimators in HRI.

% Understanding Peripersonal Space
% https://dl.acm.org/doi/10.1007/978-3-031-93718-7_17
% Biosensing

% Understanding human impressions against robots
% https://ieeexplore.ieee.org/abstract/document/8172367
% https://link.springer.com/article/10.1007/s12369-019-00601-3

% The dialog usually starts with non-committal small talk before getting to the substantial part of the conversation [8]. 

\section{Methodology}
\subsection{Participants}
We recruited 19 participants (14 female, 5 male), aged between 22 and 35 years (Mean = 25.79) from the University of Kaiserslautern-Landau, originally from a diverse range of countries, including Turkey, Iran, India, Russia, and Germany. 
We instructed participants to engage in conversation with a robot without any constraints. They answered the robot's questions about their lives and kept the conversation going while adhering to the topics offered by the robot.
%The participants' heights ranged between 156 cm and 191 cm. (Mean = 170.24 cm). 
Informed consent was obtained in compliance with GDPR regulations. The study was conducted with the approval of the Ethics Committee of the German Research Center for Artificial Intelligence (DFKI) and in accordance with the Declaration of Helsinki.

\begin{figure}[b]
  \centering
  \includegraphics[width=0.4\textwidth]{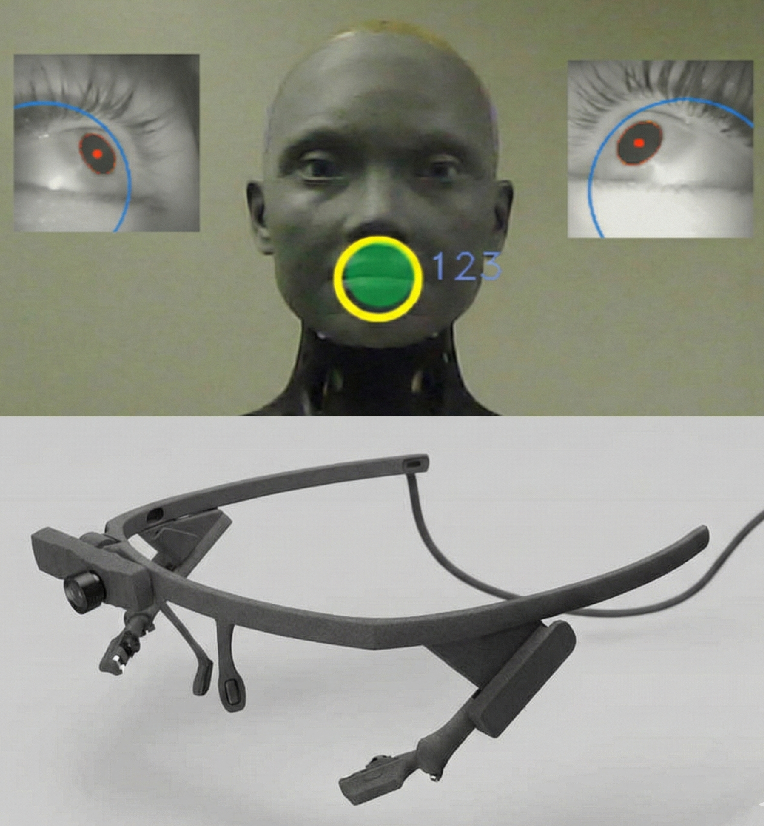}
  \caption{Ameca through Eyegaze Detection and Pupil Core Eye-Tracking Set}  
  \label{fig:ameca_and_eyegaze}
\end{figure}

\subsection{Apparatus}
\subsubsection{Robot}
Ameca, an SHR (social humanoid robot) by \citep{ameca_engineeredarts}, served as a conversational partner. Its appearance is a standing, gender-neutral human, 187 cm tall, with silicone skin on its face and hands. Its face, upper torso, arms, and hands move fluidly with 52 degrees of freedom. It operates on the EMAH robotic system, a robust framework developed at the RRLab, that enables real-time interaction. Its functionalities include gaze tracking of humans, speech recognition, generation of human-like movements and gestures during conversation, and lip syncing to speech generated by the default text-to-speech (TTS) system. 

\subsubsection{Dialogue}
The dialogue content was generated by Gemma 3~\citep{gemma3}, an open-source large language model (LLM) that ran on a PC in the experiment room and was powered by \citep{ollama}. As in ``SensPS'' \citep{watanabe_sensps_2025}, the dialogue was prompted by an inquiry about hobbies, travel, and movies. All participants engaged in conversation about the offered topics, discussing them at length. A stand-alone microphone was used for input.

\subsubsection{Eye-Tracking System}
Table~\ref{tab:pupil_features} shows the raw data features use in this study.
Fig.~\ref{fig:ameca_and_eyegaze} shows the recordings from the cameras of the eye-tracking setup, the calculated pupil positions, and eye gaze mapping. The eye-tracking setup is shown below.
Eye movements were recorded using the Pupil Core headset (Pupil Labs GmbH, \cite{kassner2014pupil}, a mobile binocular eye-tracking system. The device is equipped with two infrared (IR) eye cameras that capture pupil metrics at a sampling rate of 200 Hz. Simultaneously, a forward-facing world camera records the scene from an egocentric perspective with a resolution of 1920$\times$1080 pixels. This configuration enables the mapping of gaze onto visual stimuli within a 90-degree field of view. Weighing only 35 g, the headset is unobtrusive and suitable for continuous use. Lighting conditions were kept constant to ensure pupil data consistency.
%\subsubsection{Questionnaires}
%We recorded \textbf{individual characteristics} that may influence comfort during human-robot interaction, including gender, age, robot familiarity, country of origin, and height \citep{samarakoon2022review}.
%To define comfort on every distance a Likert scale 6-item questionnare was used, the same as in ``SensPS'' study. We administered the Likeability and Animacy subscales of the Godspeed Series in a post-interaction questionnaire. Each scale contains five semantic-differential items that are rated on a 5-point Likert scale. Participants completed these items after the interaction, providing a summary of how friendly, natural, and lifelike they perceived the robot to be. 

\subsection{Data Collection}
Upon arrival, participants provided consent and completed a demographic questionnaire. Participants were seated on a high chair to align with the standing robot. We selected a seated posture as body position significantly affects proxemic preferences \cite{samarakoon2022review}, consistent with experimental setup of ``SensPS'' involving seated participants \citep{watanabe_sensps_2025}. After calibrating the Pupil Core glasses, the experimenter initiated the robot dialogue and eye-tracking systems.

\begin{figure}[t]
  \centering
  \includegraphics[width=0.5\textwidth]{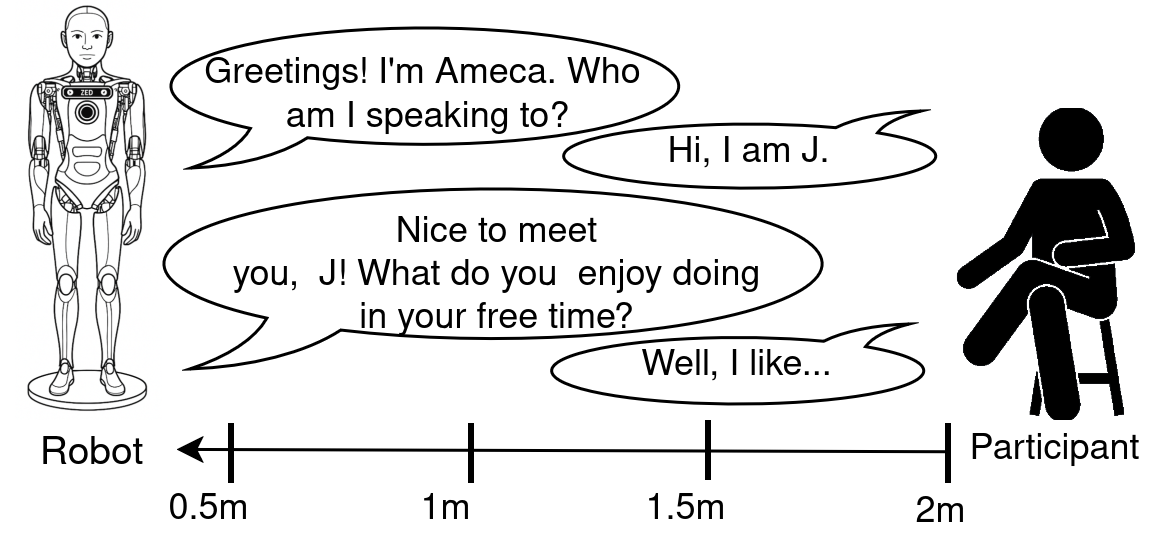}
  \caption{Experiment overview. Participants communicate with a robot within a set distance.}
  \label{fig:experiment_overview}
\end{figure}

We instructed participants to support conversation with no boundaries, aiming for an effect of a small talk, as it was found to increase trust in a robot \citep{babel_small_2021}.

\begin{figure}[h!]
  \centering
  \includegraphics[width=0.48\textwidth]{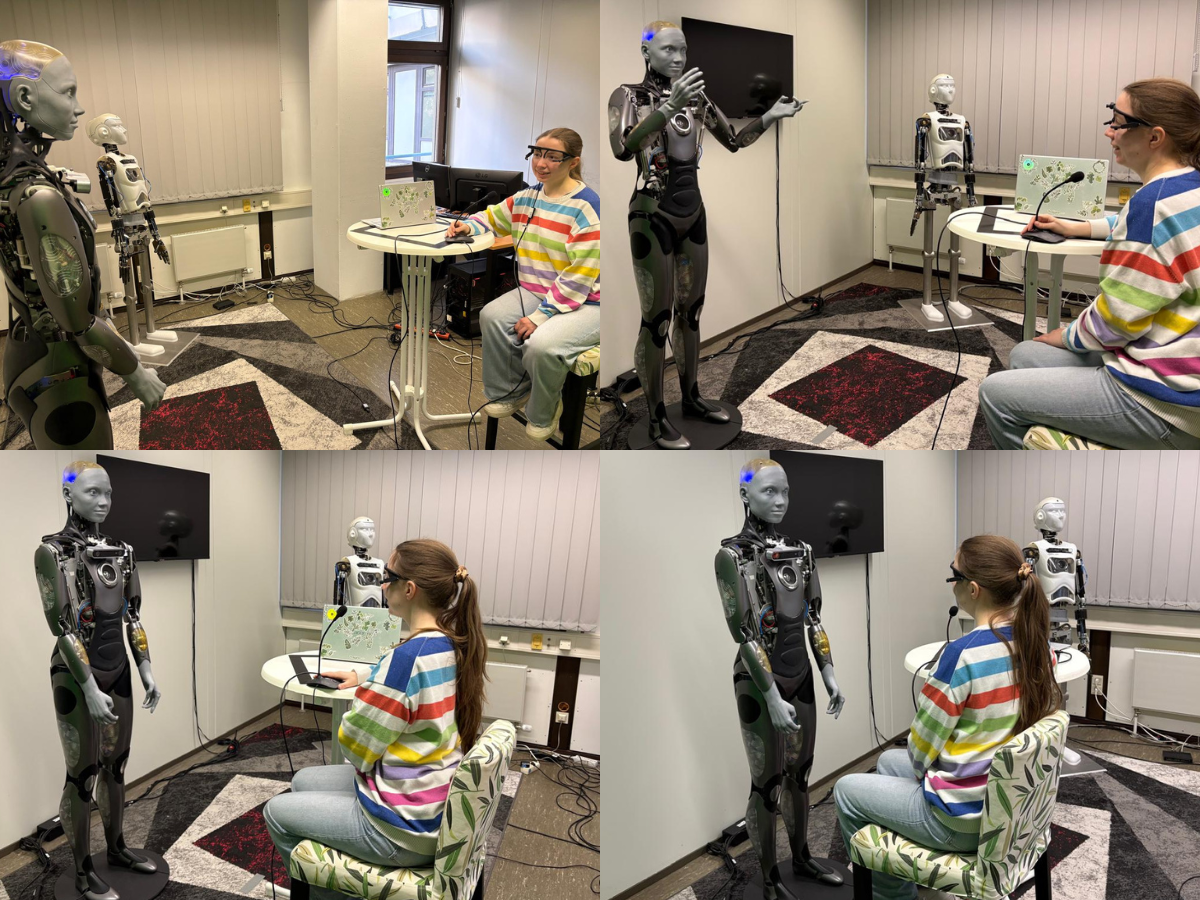}
  \caption{Experiment setup with four different distance conditions.}
  \label{fig:experimental_condition}
\end{figure}

The interaction began at a distance of 2.0 m. After the robot greeted the participant, the experimenter left the room for a three-minute interaction block. The experimenter then returned to stop the recording and administer an intermediate questionnaire. We repeated this procedure with the participant positioned incrementally closer at 1.5 m, 1.0 m, and 0.5 m (see Fig~\ref {fig:experiment_overview} and Fig~\ref {fig:experimental_condition}).

 Overall, the experimental session consisted of four three-minutes talks, with 1-minute breaks between for a questionnaire and adjusting the distance, total 12 minutes of the conversation. we instructed participants to ask the robot to repeat the last question after every repositioning. Finally, participants removed the eye-tracking equipment and completed a closing questionnaire. The average total duration of the experiment was 50 minutes.

\begin{table*}[t!]
  \centering
  \renewcommand{\arraystretch}{1.2}
  \caption{Feature list use in this study.}
  \label{tab:features_list_simple}
  \begin{tabular}{llcc}
  \toprule
  Feature & Description & Formula & Type \\
  \midrule
  min (s) & Minimum value & $\min_i(s_i)$ & T,F \\
  max (s) & Maximum value & $\max_i(s_i)$ & T,F \\
  range (s) & Difference between max and min & $\max_i(s_i)-\min_i(s_i)$ & T \\ 
  mean (s) & Mean of the signal & $\frac{1}{N}\sum_{i=1}^N s_i$ & T,F \\
  std (s) & Standard deviation & $\sqrt{\frac{1}{N}\sum_{i=1}^N (s_i-\bar{s})^2}$ & T,F \\
  mad (s) & Median absolute deviation & $median_i(|s_i - median_j(s_j)|)$ & T,F \\
  rms (s) & Root mean square & $\sqrt{\frac{1}{N}\sum_{i=1}^N s_i^2}$ & T \\
  energy (s) & Average signal energy & $\frac{1}{N}\sum_{i=1}^N s_i^2$ & T,F \\
  sma ($s_1,s_2,s_3$) & Signal magnitude area & $\frac{1}{3}\sum_{i=1}^3\sum_{j=1}^N |s_{i,j}|$ & T \\
  iqr (s) & Interquartile range & $Q3(s)-Q1(s)$ & T \\
  autoreg (s) & 4th-order Burg AR coefficients & $a = arburg(s,4),~ a\in\mathbb{R}^4$ & T \\
  correlation ($s_1,s_2$) & Pearson correlation & $\frac{C_{1,2}}{\sqrt{C_{1,1}C_{2,2}}}$ & T \\
  angle ($s_1,s_2,s_3,v$) & Angle between mean signal and vector & 
  $\tan^{-1}(\|[\bar{s}_1,\bar{s}_2,\bar{s}_3]\times v\|,\,[\bar{s}_1,\bar{s}_2,\bar{s}_3]\cdot v)$ & T \\
  skewness (s) & Signal skewness & $E\!\left[\left(\frac{s-\bar{s}}{\sigma}\right)^3\right]$ & F \\
  kurtosis (s) & Signal kurtosis & $\frac{E[(s-\bar{s})^4]}{E[(s-\bar{s})^2]^2}$ & F \\
  maxFreqInd (s) & Index of maximum frequency & $\arg\max_i(s_i)$ & F \\
  meanFreq (s) & Frequency-weighted average & $\frac{\sum_{i=1}^N i s_i}{\sum_{j=1}^N s_j}$ & F \\
  energyBand (s,a,b) & Energy in frequency band $[a,b]$ & $\frac{1}{b-a+1}\sum_{i=a}^b s_i^2$ & F \\
  psd (s) & Power spectral density & $\frac{1}{Freq}\sum_{i=1}^N s_i^2$ & F \\
  \bottomrule
  \end{tabular}
  
  \vspace{2mm}
  {\footnotesize N: signal length, Q: quartile,  T: time domain, F: frequency domain.}
\end{table*}

\subsection{Data Preprocessing}
Before analysis, raw data were extracted and processed through Pupil Player, software tools provided by Pupil Labs, to remove noise and artifacts. Three recordings (one recording is one distance for one participant) were excluded from the analysis due to insufficient eye-gaze data. First 20 seconds and last 10 seconds of every recording were excluded to account for the time an experimenter was in the room. Then, we filtered the data, keeping only those with a confidence level of at least 80\%. 

Table~\ref{tab:features_list_simple} represents the preprocessing methods use in this study, we apply use all the session data to preprocess and extract these features. The preprocessing approach refer from the study by \cite{nakamura2019waistonbelt}.

We labeled the data with binary classes, "comfortable" vs. "not comfortable". While the SensPS study \citep{watanabe_sensps_2025} employed a strict threshold of 10/10 to define comfort, we anticipated that the current experimental conditions would lower subjective scores. Consequently, we established a threshold of >=8 to define the 'comfortable' class. 

\subsection{Machine Learning and Deep Learning Models}
\subsubsection{Machine Learning}

%For our machine learning approach, we use \gls{svm}, \gls{dt}, and \gls{rf} for binary classification.

% - IFAC readers know the models, we dont have to tell them - only specifics. 

%\gls{svm} is a supervised learning technique that determines the optimal hyperplane separating two classes by maximizing the margin between them.
%\gls{dt} is a non-parametric supervised method used for both classification and regression. 
%It partitions the dataset into smaller subsets based on feature values, forming a hierarchical tree of decision rules. was configured with a maximum depth of 10 to mitigate the risk of overfitting.
%\gls{rf} is an ensemble-based algorithm that constructs multiple decision trees during training and predicts class labels by taking a majority vote. 
%This approach enhances accuracy and reduces the likelihood of overfitting.
%All three methods are applied to categorize data into two groups: comfort and discomfort.

To establish interpretable baselines for comfort detection, we employed \gls{svm}, \gls{dt}, and \gls{rf} classifiers. 
The \gls{dt} model was explicitly configured with a maximum depth of 10 to balance model complexity with the risk of overfitting, while the \gls{rf} was implemented to reduce variance through ensemble averaging.

To evaluate model generalization across unseen subjects, we adopted a Leave-One-Subject-Out (LOSO) cross-validation protocol. In each fold, one participant was isolated as the test set, while the remaining participants were in the training set. To address class imbalance in the training phase, we applied random undersampling to the majority class (``not comfortable'') for a strictly balanced 1:1 distribution. The test data for the held-out subject remained unmodified to preserve the real-world class distribution

\subsubsection{Deep Learning Architectures}
We extended our analysis to deep learning architectures to capture complex, high-dimensional interactions within the physiological signals. We implemented Convolutional Neural Networks (CNNs), specifically VGG16 \citep{simonyan2014very} and the MobileNet family (V1, V2, V3) \citep{howard2017mobilenets}. While VGG16 provides a deep feature extraction capability, the MobileNet architectures were selected for their computational efficiency, which is particularly relevant for deployment on wearable computing platforms.

Furthermore, we utilized a Transformer architecture \citep{vaswani2017attention} incorporating multi-head self-attention mechanisms. This model was chosen to explicitly capture long-range temporal dependencies in the gaze data sequence, following the success of this architecture in human-human proxemics studies.

\section{Result and Discussion}

\begin{table}[h!]
\centering
\caption{Descriptive statistics for comfort ratings at four robot--human distances.}
\label{tab:comfort_descriptives}
\begin{tabular}{lcccc}
\toprule
 & \multicolumn{4}{c}{Distance} \\
\cmidrule(lr){2-5}
Statistic & 2\,m & 1.5\,m & 1\,m & 0.5\,m \\
\midrule
Mean            & 8.32 & 7.71 & 7.74 & 7.64 \\
Std.\ Deviation & 1.77 & 1.99 & 1.96 & 2.07 \\
Minimum         & 2.50 & 2.00 & 2.83 & 2.83 \\
Maximum         & 10.00 & 10.00 & 10.00 & 9.83 \\
\bottomrule
\end{tabular}
\vspace{-2mm}
\end{table}

We analyzed comfort ratings using a one-factor repeated-measures ANOVA. The analysis revealed a significant main effect of Distance, $F(3,54) = 3.25$, $p = .029$, $\eta^{2}_{p} = .15$, confirming that proximity influences perceived comfort. Pairwise comparisons with Bonferroni correction did not reach the corrected significance threshold, likely due to the limited sample size and high inter-subject variability (Table \ref{tab:comfort_descriptives}).

\begin{figure}[t!]
  \centering
  \includegraphics[width=0.48\textwidth]{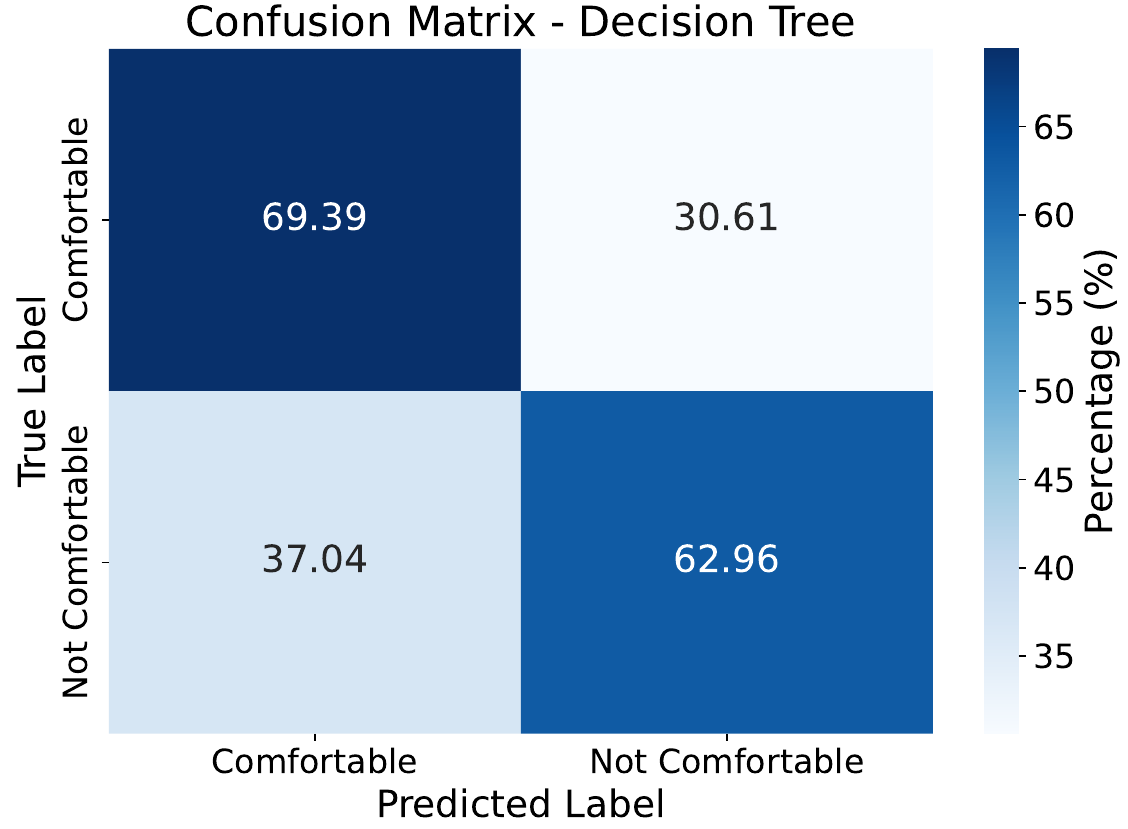}
  \caption{Confusion matrix for Decision Tree model.}  
  \label{fig:dt_confusion}
\end{figure}

Table \ref{tab:model_comparison} summarizes the comparative performance of the implemented models. Decision Tree classifier achieved a macro-averaged F1 score of 0.73, outperforming all other models, including Random Forest (F1 = 0.61) and Transformer (F1 = 0.57). Figure \ref{fig:dt_confusion} shows a confusion matrix of the Decision Tree model, with comfort being predicted with higher accuracy than discomfort. 
Given the variability of physiological data and Leave-One-Subject-Out evaluation, this result indicates that the Decision Tree generalizes effectively to unseen subjects. It suggests that comfort is defined by specific, hierarchical threshold values rather than complex, high-dimensional interactions.

These findings stand in contrast to the results reported in the "SensPS" study \citep{watanabe_sensps_2025}, where the Transformer model achieved the highest performance (F1=0.87) using a smaller cohort of 10 participants, significantly outperforming the Decision Tree (F1=0.54) and Random Forest (F1=0.67). While the overall performance of almost all models in our study is lower than in ``SensPS'', Decision Tree performed significantly higher (F1 = 0.73 vs 0.54). In our study, which included a larger and likely more heterogeneous group of 19 participants , the Transformer failed to generalize (F1 = 0.57). Better performance of the Decision Tree (F1 = 0.73) in our study suggests that as inter-subject variability increases, simple, interpretable rules based on robust physiological thresholds become more effective than complex deep learning representations, which may overfit to individual characteristics.

\begin{table}[t!]
\centering
\caption{Classification performance comparison: Accuracy, Precision, Recall, and F1 Score.}
\label{tab:model_comparison}
\begin{tabular}{lcccc}
\toprule
Model       & Accuracy & Precision & Recall & F1 Score \\
\midrule
SVM         & 0.5395 & 0.5553 & 0.5597 & 0.5355 \\
\textbf{DT} & \textbf{0.671} & \textbf{0.773} & \textbf{0.694} & \textbf{0.731} \\
RF          & 0.5395 & 0.6667 & 0.5714 & 0.6154 \\
VGG16       & 0.5658 & 0.6818 & 0.6122 & 0.6452 \\
MN          & 0.5000 & 0.6486 & 0.4898 & 0.5581 \\
MNV2        & 0.5132 & 0.6364 & 0.5714 & 0.6022 \\
MNV3        & 0.4737 & 0.5957 & 0.5714 & 0.5833 \\
Transformer & 0.5263 & 0.6857 & 0.4898 & 0.5714 \\
\bottomrule
\end{tabular}
\end{table}

\begin{figure*}[h!]
  \centering
  \includegraphics[width=\textwidth]{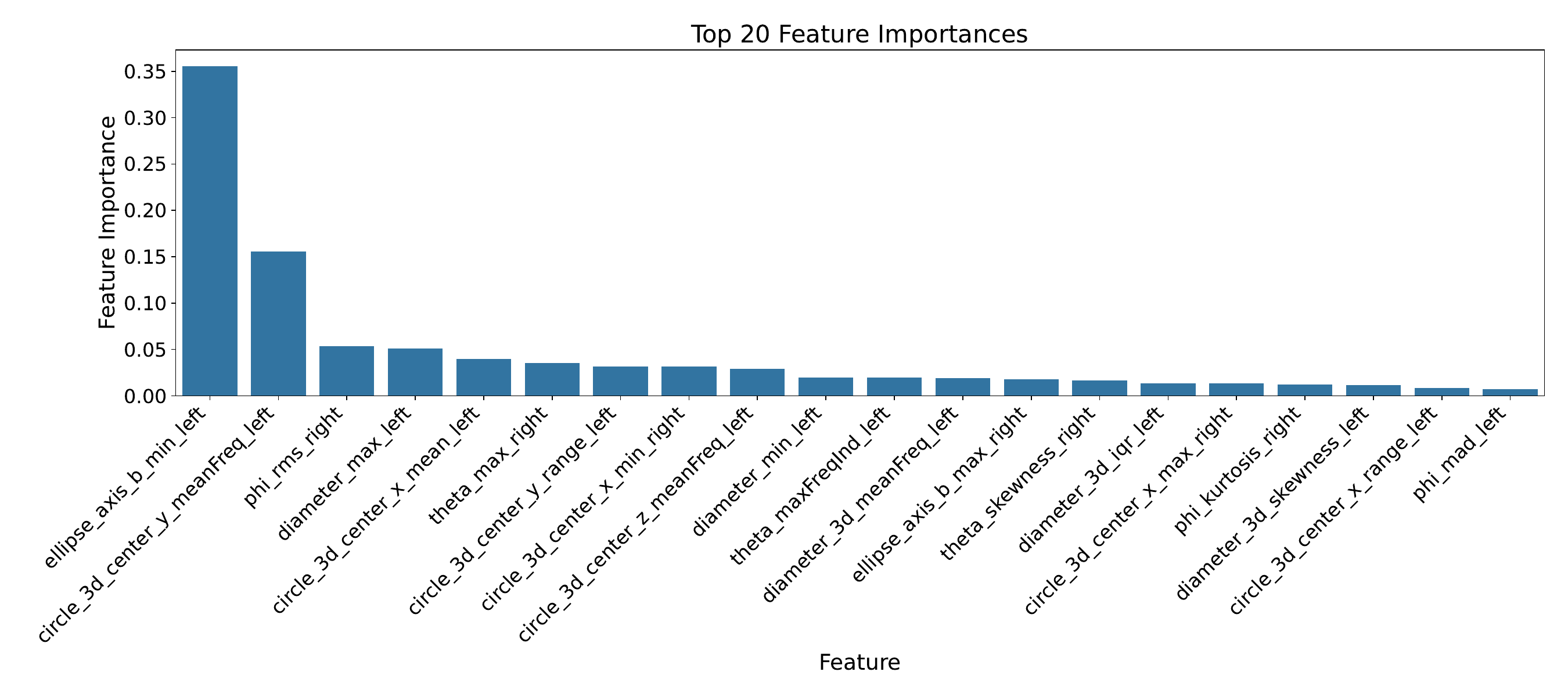}
  \caption{Feature importance for Decision Tree model.}  
  \label{fig:dt_feature_importance}
\end{figure*}

Fig. \ref{fig:dt_feature_importance} shows the feature importance calculated via Gini impurity. Feature with the biggest weight (= 0.35) represents the minimal pupil diameter. This aligns with the link between the Autonomic Nervous System (ANS) and arousal; pupil constriction correlates with parasympathetic dominance and rest, relaxation, and lower cognitive load \cite{steinhauer2022pupils}.
In contrast, the most  important (= 0.35) feature in ``SensPS'' represents skewness of a gaze point within their optimal Random Forest model, implying that the difference in experiment setup might elicit different eye gaze behavior.

\section{Limitation and Future Work}

We acknowledge several limitations. First, the use of wearable eye-tracking glasses (Pupil Core) introduces an experimental artifact that may impact ecological validity. The physical sensation of the headset and the need for calibration breaks during the interaction flow may heighten user awareness of the observation, potentially altering natural comfort responses. Second, while the robot is highly anthropomorphic, it remains immobile, which simplifies the proxemic dynamics compared to a moving human or mobile robot.

In future, intrusive eye tracking systems can be replaced with remote gaze estimation systems \cite{bhatt_appearance-based_2024}. Validating our findings with camera-based tracking will enable the development of non-invasive, adaptive robotic systems capable of modulating distance in real-time based on user comfort.

\section{Conclusion}

In summary, we demonstrated that gaze features can serve as objective estimators of user comfort during interactions with a social humanoid robot. We developed an interpretable Machine Learning pipeline, with a Decision Tree model achieving an F1-score of 0.73. Our findings indicate that minimal pupil diameter is the most robust predictor of comfort in this context. These insights provide a foundation for designing proxemic-aware robots that can implicitly detect and adjust to user discomfort.

% \begin{ack}
% Place acknowledgments here.
% \end{ack}

\bibliography{main}

\appendix
% \section{A summary of Latin grammar}    % Each appendix must have a short title.
% \section{Some Latin vocabulary}              % Sections and subsections are supported in the appendices.
\end{document}